%% file: main.tex
\documentclass[10pt,twocolumn,letterpaper]{article}

\usepackage[pagenumbers]{cvpr} 

\usepackage{graphicx}
\usepackage{amsmath}
\usepackage{amssymb}
\usepackage{booktabs,siunitx,caption}
\usepackage{algorithm}
\usepackage{algpseudocode}
\usepackage{pifont}
\usepackage{multirow}
\usepackage{makecell}
\usepackage{listings}

\usepackage{etoolbox}
\makeatletter
\AfterEndEnvironment{algorithm}{\let\@algcomment\relax}
\AtEndEnvironment{algorithm}{\kern2pt\hrule\relax\vskip3pt\@algcomment}
\let\@algcomment\relax
\newcommand\algcomment[1]{\def\@algcomment{\footnotesize#1}}
\renewcommand\fs@ruled{\def\@fs@cfont{\bfseries}\let\@fs@capt\floatc@ruled
  \def\@fs@pre{\hrule height.8pt depth0pt \kern2pt}%
  \def\@fs@post{}%
  \def\@fs@mid{\kern2pt\hrule\kern2pt}%
  \let\@fs@iftopcapt\iftrue}
\makeatother

\algdef{SE}[SUBALG]{Indent}{EndIndent}{}{\algorithmicend\ }%
\algtext*{Indent}
\algtext*{EndIndent}

\newcommand{\xhdr}[1]{\vspace{5pt} \noindent {\textbf{#1}}}

\newcommand{\ModelName}{Stochastic Backpropagation}
\newcommand{\ModelAbbr}{SBP}

\usepackage[pagebackref,breaklinks,colorlinks]{hyperref}

\usepackage[capitalize]{cleveref}
\crefname{section}{Sec.}{Secs.}
\Crefname{section}{Section}{Sections}
\Crefname{table}{Table}{Tables}
\crefname{table}{Tab.}{Tabs.}

\def\cvprPaperID{***} 
\def\confName{CVPR}
\def\confYear{2022}

\begin{document}

\title{\ModelName: \\ A Memory Efficient Strategy for Training Video Models}

\author{
    Feng Cheng$^1$\thanks{The work was done during an Amazon internship.} \hspace{0.35cm} Mingze Xu$^2$\thanks{Corresponding Author.} \hspace{0.25cm} Yuanjun Xiong$^2$ \hspace{0.25cm} Hao Chen$^2$ \hspace{0.25cm} Xinyu Li$^2$ \hspace{0.25cm} Wei Li$^2$ \hspace{0.25cm} Wei Xia$^2$ \\ [.5ex] $^1$UNC Chapel Hill \hspace{0.9cm} $^2$AWS AI Labs \\ [.5ex]
    {\tt\small fengchan@cs.unc.edu, \{xumingze,yuanjx,hxen,xxnl,wayl,wxia\}@amazon.com}
}

\maketitle

\begin{abstract}
    We propose a memory efficient method, named~\ModelName~(\ModelAbbr), for training deep neural networks on videos. It is based on the finding that gradients from incomplete execution for backpropagation can still effectively train the models with minimal accuracy loss, which attributes to the high redundancy of video. \ModelAbbr~keeps all forward paths but randomly and independently removes the backward paths for each network layer in each training step. It reduces the GPU memory cost by eliminating the need to cache activation values corresponding to the dropped backward paths, whose amount can be controlled by an adjustable keep-ratio. Experiments show that \ModelAbbr~can be applied to a wide range of models for video tasks, leading to up to 80.0\% GPU memory saving and 10\% training speedup with less than 1\% accuracy drop on action recognition and temporal action detection.
\end{abstract}

\input{introduction}
\input{related_work}
\input{our_approach}
\input{experiments}

\input{conclusion}

{\small
\bibliographystyle{ieee_fullname}
\bibliography{egbib}
}

\end{document}

%% file: introduction.tex
\section{Introduction}
\label{sec:intro}

One of the common challenge in training video understanding models is the limited availability of GPU memory.
Although video models, such as those for action recognition~\cite{simonyan2014two,tran2015learning,wang2016temporal,bertasius2021space} and temporal action detection~\cite{caba2015activitynet,de2016online}, are known to benefit from capturing context features over the entire span of a video~\cite{wu2019long,wang2016temporal},
it is not always feasible to end-to-end train them while taking as many input frames (\eg, typically hundreds) as they need.
For example, when training a temporal action detector using ResNet-50~\cite{he2016deep} as feature extractor, 128 frames as inputs, and batch size 4,
the feature extractor itself costs \textit{over} 40 GB memory that could exceed the limit of most modern GPUs.

\begin{figure}[t]
    \centering
    \includegraphics[width=\linewidth]{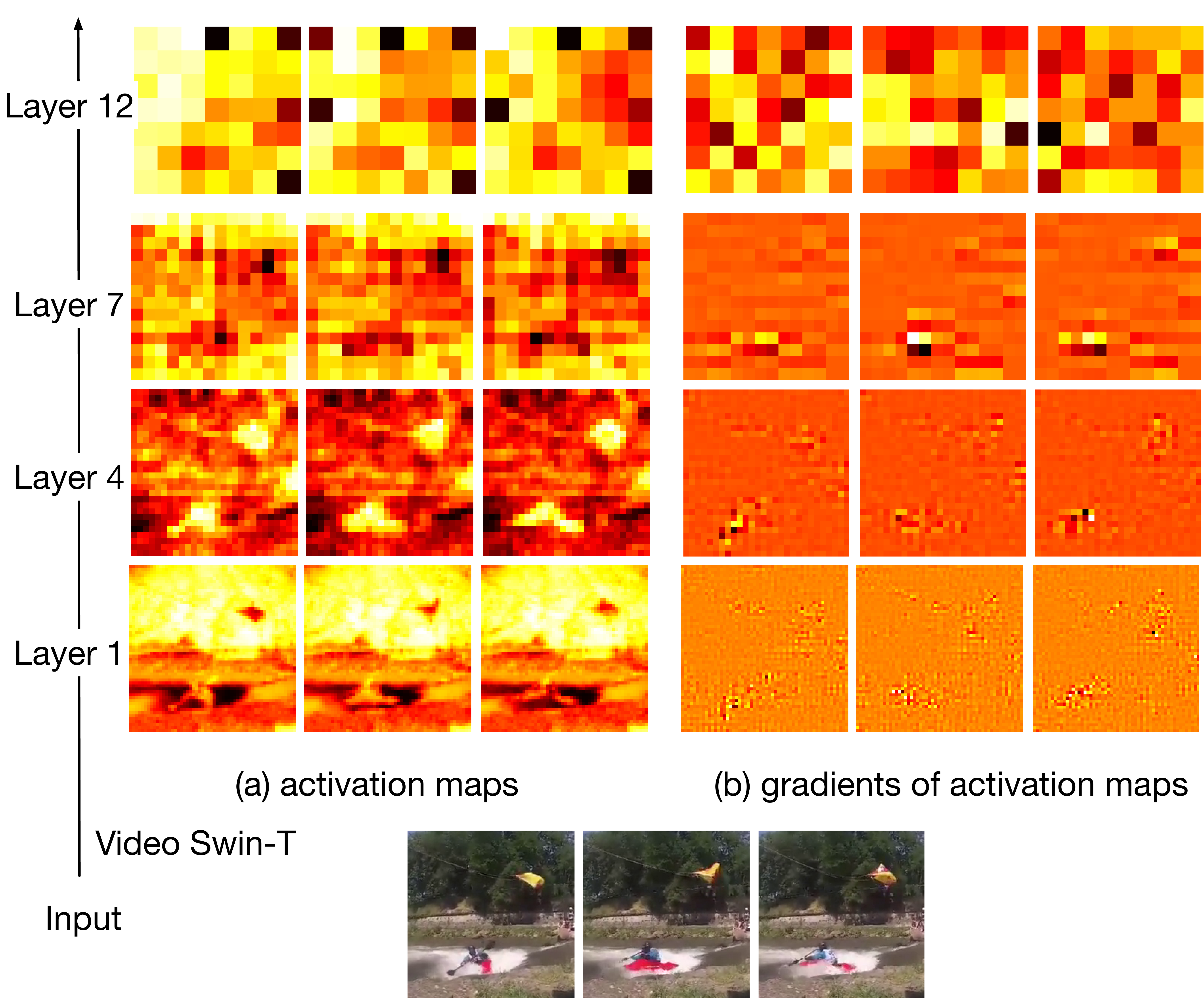}
    \caption{The activations and gradients are obtained from different layers (\eg, layer 1, 4, 7, and 12) of Video Swin-T~\cite{liu2021video}. The activations and gradients between frames are similar at the bottom layers and are quite different at the top layers, which matches our assumption that there is much redundancy between the frames at the bottom layers but less at the top layers.}
    \label{fig:redundancy}
\end{figure}

On the other hand, video data is highly redundant, which suggests opportunity for optimization. We observe that in video models, the activation maps and gradients of different frames are similar at the bottom layers but become more and more different at the top layers.
An example of Video Swin-T~\cite{liu2021video} on Kinetics-400~\cite{carreira2017quo} dataset is shown in Fig.~\ref{fig:redundancy}.
During model training, the tiny difference in the seemly similar features (\ie, fine-grained information) of the bottom layers is crucial for producing the difference in the top layers, however, they may not make much difference in terms of updating the parameters of these bottom layers.
Thus, we hypothesize that complete computation is necessary for the activation maps (forward paths) to extract important semantic information but could be \emph{unnecessary} for the gradients (backward paths).

In this paper, we propose a memory saving technique, named~\textit{~\ModelName~(\ModelAbbr)}. 
Different from previous work that jointly removes the forward and backward paths~\cite{akbari2021vatt},
\ModelAbbr~randomly drops a proportion of the individual backward paths in backpropagation while keeping all the forward paths.
Furthermore, we found that this random removal of backward paths can happen in a layer-wise manner, where we remove immediately connected paths of certain randomly sampled neurons in each layer.
Note that, due to the chained dependency in the backpropagation algorithm, this would lead to incompletely computed gradients for all network parameters. However, we empirically found that gradients from this incomplete computation could still be sufficient for effectively updating the network parameters, as long as the overall computation graph is preserved.
This makes \ModelAbbr~easy to be implemented without the need to account for the change of dependency across multiple layers.
The memory saving of \ModelAbbr~is from the avoidance of caching the corresponding activation maps on removed computation paths.
Since the memory can be safely re-used in backpropagation, the backward process uses almost no (one layer at most) additional memory\footnote{This mechanism is already encapsulated in most modern deep learning frameworks, such as PyTorch and Tensorflow.}, which makes the activation caching the major source of memory cost during training.
Thus, the saving amount solely depends on the keep-ratio and can be significant when the keep-ratio is low.

\ModelAbbr~is a general technique that can be applied to a wide range of video tasks and models.
Experimental results show that training with~\ModelAbbr~uses only $0.2\times$$\sim$$0.5\times$ the GPU memory and speedup $1.1\times$$\sim$$1.2\times$ the training with less than $1\%$ loss of accuracy on Kinetics-400~\cite{carreira2017quo} and Epic-Kitchen-55~\cite{damen2018scaling} for action recognition and less than $1\%$ loss of mAP on THUMOS'14~\cite{THUMOS14} for temporal action detection.

\begin{figure}[t]
    \centering
    \includegraphics[width=\linewidth]{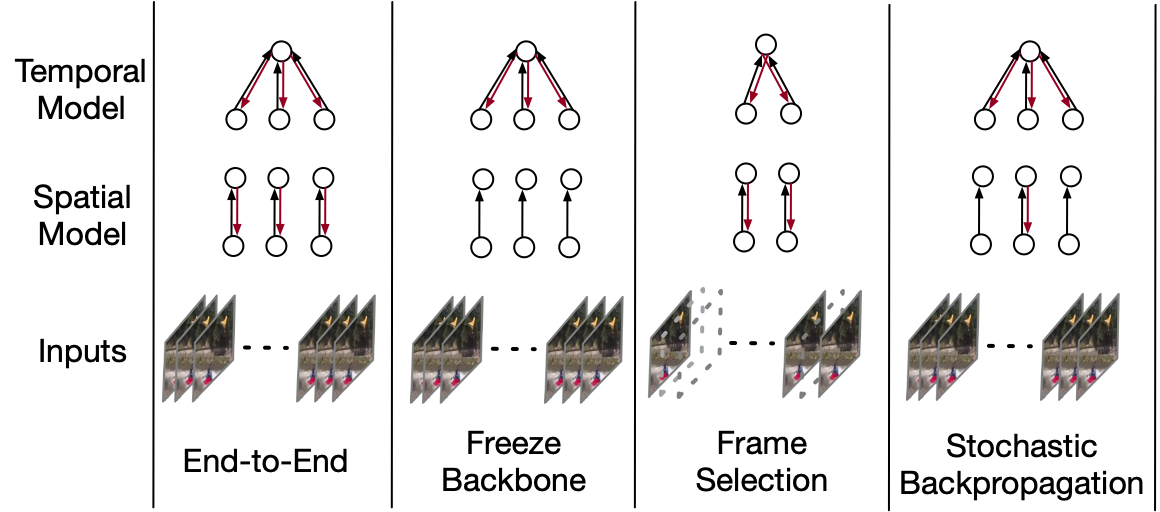}
    \vspace{-4mm}
    \caption{Comparison of different memory saving techniques. Different from freezing backbone or frame selection, \ModelName~randomly removes the backward paths when training the spatial model while keep them for the temporal model.}
    \label{fig:method_comparison}
\end{figure}

%% file: related_work.tex
\section{Related Work}

\vspace{-5pt}
\xhdr{Video Understanding}.
Action recognition~\cite{simonyan2014two,tran2015learning,wang2016temporal,carreira2017quo,tran2018closer} aims to classify the activities of one or more agents in videos of various types and applications, from privacy-sensitive~\cite{xu2018fully} and surveillance cameras~\cite{sultani2018real} to egocentric videos~\cite{li2015delving,ma2016going,damen2018scaling}.
Given an untrimmed video, temporal action localization estimates the start and end times of each action instance.
One common practice~\cite{dai2017temporal,buch2017sst,gao2017turn,chao2018rethinking,lin2018bsn,lin2019bmn,xu2020g} is to first generate action proposals, then identify their action classes and temporal boundaries.
On the other hand, the ''bottom-up'' fashion~\cite{shou2017cdc,zhao2017iccv} first makes frame-level dense predictions and groups them as action instances.
Online action detection~\cite{de2016online,shou2018online,xu2019temporal,gao2019startnet,eun2020learning,qu2020lap,zhao2020privileged,gao2020woad,xu2021long}, however, recognizes actions as soon as each video frame arrives without accessing any future information.
Due to the high cost of GPU memory, most above offline and online methods cannot be trained in end-to-end manner and thus are built upon the extracted features from raw videos.
Transformers recently achieve convincing results in video tasks, such as action recognition~\cite{li2021vidtr,bertasius2021space,arnab2021vivit,liu2021video} and detection~\cite{nawhal2021activity,tan2021relaxed,xu2021long}.
However, the high computational cost and memory demands also limit most of them to be only trained on short video clips. A few methods~\cite{dai2019transformer,burtsev2020memory} focus on designing Transformers to model long-form inputs, but mainly for text.

\xhdr{Memory Saving Techniques}.
The trade-off between memory and accuracy has been a standing topic in deep learning research.
Gradient checkpoint~\cite{chen2016training} and accumulation, can save a large amount memory ($\sim$50\%) but is likely to slow down the training process.
Sparse Network~\cite{dettmers2019sparse} is applied to image recognition models but can only save the memory theoretically.
Sideways~\cite{malinowski2020sideways,malinowski2021gradient} reduce the memory cost by overwriting activations whenever new ones become available but can only be applied to causal models.
Regarding video specific methods, a popular paradigm for training temporal action detectors is to build the model upon pre-extracted features for temporal modeling and reasoning (``Freeze Backbone" in Fig.~\ref{fig:method_comparison}).
Two recent methods, AFSD~\cite{lin2021learning} and DaoTAD~\cite{wang2021rgb}, are end-to-end trained (DaoTAD also freezes the first two stages of their backbone), but they need to reduce the spatial resolution (\ie $96\times96$ in AFSD and $112\times112$ in DaoTAD).
However, freezing the backbone or reducing the input size can decrease the accuracy and is not an optimal solution.
Since video usually contains high redundancy between adjacent frames,
several work also tries to reduce the number of input frames, including frame dropout~\cite{akbari2021vatt}, early pooling~\cite{liu2021no} and frame selections~\cite{gowda2020smart} (``Frame Selection" in Fig.~\ref{fig:method_comparison}).
However, aggressively subsampling of input frames will unavoidably lose important fine-scale
information in the temporal domain, thus have a negative impact on the final accuracy~\cite{wu2019long,xu2021long}. In addition, these methods cannot be applied to tasks that require per-frame predictions (\eg, online action detection) or involve high-motion (\eg, lip-reading).
The comparisons between~\ModelAbbr~and other memory saving techniques are shown in Fig.~\ref{fig:method_comparison}

%% file: our_approach.tex
\begin{figure*}[t]
     \centering
     \begin{subfigure}[b]{0.45\textwidth}
         \centering
         \includegraphics[width=\textwidth]{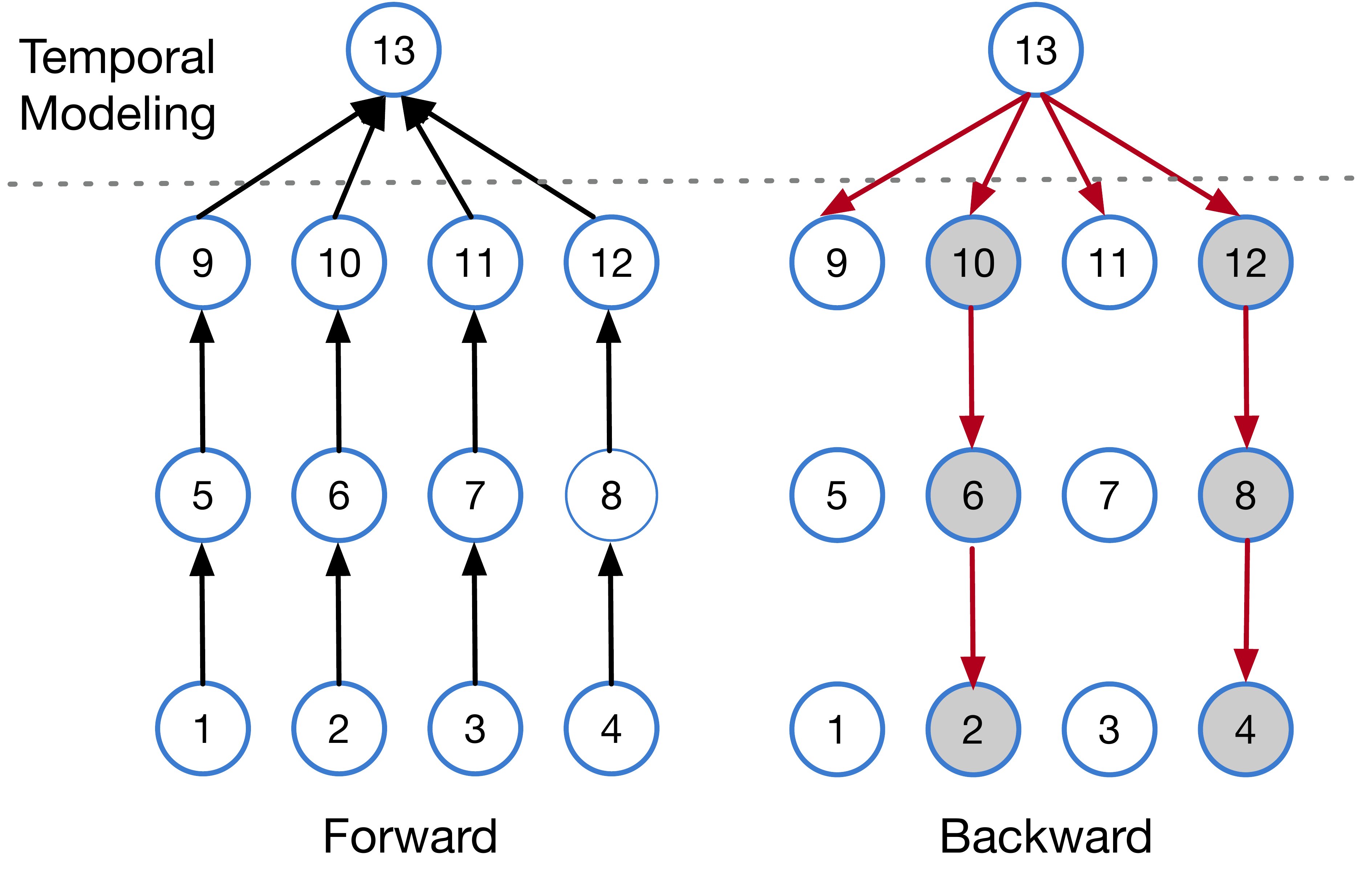}
         \caption{\ModelAbbr~on a tree model.}
          \label{fig:sbp_general_tree_model}
     \end{subfigure}
     \hfill
     \begin{subfigure}[b]{0.45\textwidth}
         \centering
         \includegraphics[width=\textwidth]{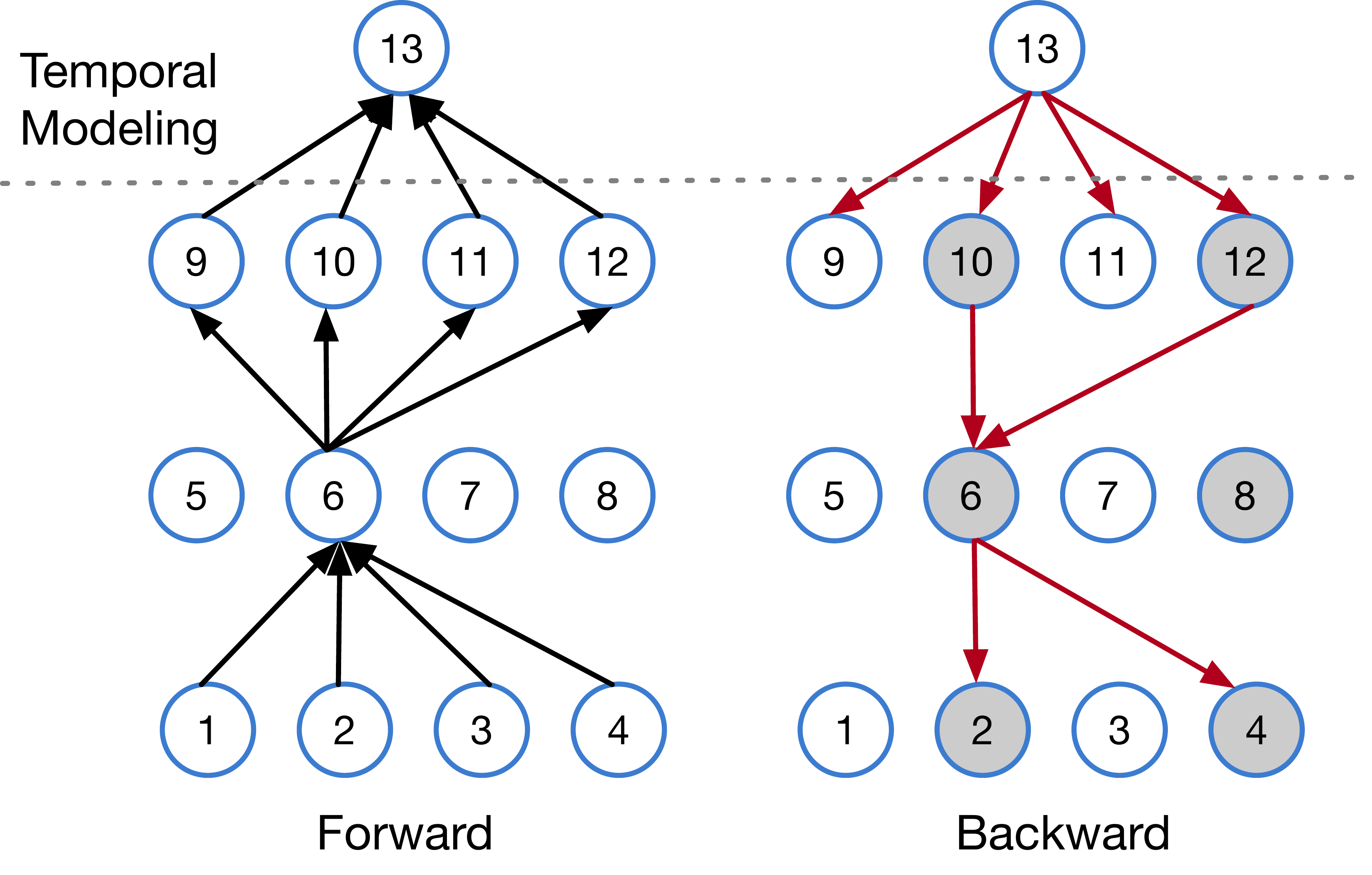}
         \caption{\ModelAbbr~on a graph model.}
          \label{fig:sbp_general_graph_model}
     \end{subfigure}
     \vspace{-1.5mm}
     \caption{Illustration of \ModelAbbr~on tree and graph models. In tree model (a), each node is only connected to the nodes with the same location in adjacent layers. In graph model (b), each node is potentially connected to all the nodes in adjacent layers (only connections of Node-6 are plotted for better visualization). During backward, the gradients will only propagate between the sampled nodes (in color gray).}
     \label{fig:sbp_model}
\end{figure*}

\section{\ModelName}
\label{sec:our_approach}

We propose a memory efficient strategy,~\textit{\ModelName~(\ModelAbbr)}, for training video models. Due to the high-redundancy at the bottom layers, \ModelAbbr~randomly drops a proportion of their gradients during backpropagation. This can save the memory since we no longer need to keep track of some of the activations that are used to calculate the gradients. In this section, we first introduce the general idea of how \ModelAbbr~works in tree and graph models that are widely used in video tasks, then explain the concrete implementations for these two types of models.

\subsection{Overview}
\label{sec:our_approach:overview}

When a model is defined, its backward computational graph is a Directed Acyclic Graph (DAG), $G(V, E)$, in which the vertices $V$ are the activations cached for gradients calculation and edges $E$ are the operations to calculate the gradients. As the model is a stack of layers, we also organize $V$ in a layer-wise fashion, that is, $V = \{V^{l_0}, V^{l_1}, V^{l_2}, \cdots, V^{l_n}\}$, in which $V^{l_i}$ is the output of layer $l_i$ and $V^{l_0}$ is the model input. We view $V^{l_i}$ as a set of feature vectors. For example, if $V^{l_i} \in \mathbb{R}^{C \times D \times H \times W}$, we view $V^{l_i}$ as $D \times H \times W$ vertices, where each vertex (we call it as \textit{node} in this paper) is a feature vector $\in \mathbb{R}^C$. 

As shown in Fig.~\ref{fig:sbp_model}, we generally divide the entire model into two parts: 1) spatial model (below dashed line) and 2) temporal model (above dashed line).
The top several layers of the model can be viewed as the temporal model as these layers mainly learn global semantic information from all the frames. All the other bottom layers are viewed as the spatial model that mainly learns the spatial and local context information.
\ModelAbbr~is only applied to the spatial model.

The tree-model is a special type of models with tree-like DAGs, whose children node is usually only connected with one father node. It is one of the simplest model structures but is widely adopted in long-term video tasks.
To better illustrate \ModelAbbr,~we first introduce \ModelAbbr~on tree-models and then extend it to more general graph-models.

\vspace{-1.0mm}
\subsubsection{\ModelAbbr~on Tree-Models}
\vspace{-1.0mm}

In the tree-models, each node is only connected to the nodes with the same location at the neighboring layers (as shown in Fig.~\ref{fig:sbp_general_tree_model}). In long-term video tasks, most methods~\cite{xu2019temporal,xu2021long,xu2020g} use a spatial backbone, such as ResNet~\cite{he2016deep}, to extract frame features. Such models can be viewed as tree-models as the frames are independently processed with each other.

Fig.~\ref{fig:sbp_general_tree_model} shows how \ModelAbbr~is done on tree-models. The several top layers are used to do temporal modeling, capturing temporal information from multiple frames. 
As there are more redundancy at the bottom layers than the top layers, we only drop the gradients at the bottom layers and keep all the gradients at the top layers. 
During normal end-to-end training, most memory is consumed at the bottom layers due to the much higher feature dimensions.
\ModelAbbr~drops gradients at the bottom layers and eliminates the necessity of tracking the corresponding intermediate nodes, which can save lots of memory.

\vspace{-1.5mm}
\subsubsection{\ModelAbbr~on Graph-Models}
\vspace{-1.0mm}

In a more general graph-model, each node is potentially connected to all the nodes at the neighboring layers (as shown in Fig.~\ref{fig:sbp_general_graph_model}). Vision Transformers~\cite{liu2021video,bertasius2021space,dosovitskiy2020image} are one type of fully-connected graph-models. Fig.~\ref{fig:sbp_general_graph_model} shows how \ModelAbbr~is done on graph-models. During forward, each node will aggregate data from all the nodes in the previous layer (only the connections of Node-6 is plotted for better visualization). During backward, similar to tree-models, the gradients are only propagated between the sampled nodes.

In tree-models, the gradients of sampled nodes are exactly the same as training using all the nodes. For example, in Fig.~\ref{fig:sbp_general_tree_model} the gradients we calculated for Node-6 are the same no matter Node-9,11,5,7 are sampled or not. But in general graph-models, as all the top nodes will propagate gradients to the bottom nodes, the gradients' calculation for the sampled nodes are not exact and will be more and more inaccurate from top to bottom. However, this will not be a problem as there are more and more redundancy from top to bottom layers. Keeping the sampled nodes the same across layers will also alleviate this problem.
We set the a uniform keep-ratio of \ModelAbbr~for all layers to simplify parameter identification. But we do note that a systematic strategy can be developed for setting layer-specific keep-ratios.

\begin{figure}[t]
    \centering
    \includegraphics[width=0.9\linewidth]{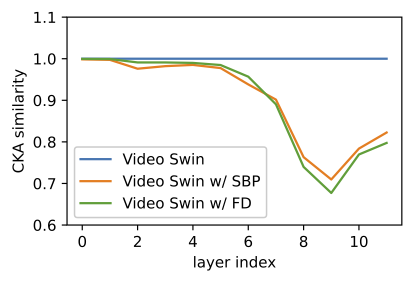}
    \vspace{-2.5mm}
    \caption{The CKA similarity~\cite{kornblith2019similarity} between \ModelAbbr,~Frame Dropout (FD)~\cite{akbari2021vatt}, and End-to-End training of the activations at the same layer on Video Swin-T~\cite{liu2021video}. The similarities at bottom layers are quite high, which indicates drop gradients at bottom layers will not hurt the model due to the high-redundancy. w/ \ModelAbbr~has higher similarity than w/ FD at the top layers, which shows the importance of keeping all gradients of top layers.}
    \label{fig:cka_similarity}
    \vspace{-2.5mm}
\end{figure}

\xhdr{Why can we effectively learn with partial gradients?} 
We believe it is because we remove the redundancy between frames but keep all important information. As shown in Fig.~\ref{fig:redundancy}, as both the activation maps and the gradients of these activation maps are very similar between the frames at the bottom layers, dropping the gradients of some frames will not hurt the model too much. 
Fig.~\ref{fig:cka_similarity} also supports this point. The CKA similarity~\cite{kornblith2019similarity} $\in [0,1]$ are used to measure the similarity of the representations between two neural networks. Frame Dropout (FD)~\cite{akbari2021vatt} is trained using only $1/4$ of the frames as in the baseline, which is equivalent to dropping $3/4$ of the gradients in all the layers. The curve of FD is the CKA similarity between the model trained using 1/4 of the frames and using all the frames (baseline). The similarity is very high ($>$0.9) for the first 6 layers, which means dropping the gradients will not degrade model's performance dramatically.

\xhdr{Comparison with frame selection methods.}
The frame selection methods~\cite{gowda2020smart,korbar2019scsampler} use a light-weight network to sample the frames from the video and then feed the sampled frames to the target model. 
Suppose we use the same frame sampler, the differences between \ModelAbbr~and frame selection methods are whether to preserve forward-path for the unsampled frames.
We argue that keeping all the forward-paths is important.
The top layers that do temporal modeling learn global temporal information from all the frames and thus the learned features contain less redundancy as shown in Fig.~\ref{fig:redundancy}. 
Dropping the forward-path as in frame-selection methods will impact the temporal modeling as the less redundant temporal model can only learn from a sampled set of frames.
As shown in Fig.~\ref{fig:cka_similarity}, keeping the gradients (with \ModelAbbr)~at the top layers has higher CKA similarity than dropping the gradients (with FD).

\xhdr{Comparison with Gradient Dropout~\cite{tseng2020regularizing}.}
Both \ModelAbbr~and Gradient Dropout drop the gradients during backward. 
Our \ModelAbbr~aims to save the memory by dropping a large proportion (\ie, 75\%) of the backward paths while Gradient Dropout aims to regularize the training by randomly zero-out a small proportion (10\%$\sim$20\%) of gradients.
We explore the application of gradient dropout in a new direction and then design a strategy to drop the backward paths that can achieve practical memory savings.

\subsection{Instantiations}
\label{sec: sbp_stmodel}

Tree-models are typically adopted in long-term video tasks that take hundreds of frames as inputs. 
For short-term video tasks, such as action recognition, the fully-connected graph-models (\eg, Video Transformers~\cite{arnab2021vivit,liu2021video,li2021vidtr,bertasius2021space}) achieve better accuracy.
In this section, we instantiate on these two types of models.

\vspace{-3.5mm}
\subsubsection{Spatial-then-Temporal (StT) Models}
\vspace{-1.5mm}

For long-term tasks such as online action detection and offline action detection, most existing methods are Spatial-then-Temporal models (StT models).
StT models are instances of tree-models (Fig.~\ref{fig:sbp_general_tree_model}). They first use a spatial model (2D- or 3D-CNN using sliding windows) to extract the spatial features for each frame / chunk and then feed the extracted features to a temporal model (RNN, Transformer, \etc) for temporal modeling.

Denote the spatial model as $f_s$ and temporal model as $f_t$, the input video $x \in \mathbb{R}^{C \times D \times H \times W}$ is a clip of $D$ frames with spatial resolution $H \times W$. We omit the batch dimension for simplicity. The spatial model takes input of size $\mathbb{R}^{C \times K \times H \times W}$, in which $K$ is the chunk size. For example, 2D-CNN takes a single frame as input ($K=1$) and 3D-CNN  takes a chunk of $K$ frames as input. So the forward pass of this model would be:
\begin{align}
    x_i &= x[\, :, i*K : (i+1)*K], i \in \{0, \cdots, n\}\\
    h   &= \{f_s(x_0), f_s(x_1),\cdots,f_s(x_n)\} , n = T/K - 1\\
    y   &= f_t(h) 
\end{align}
In this model, the spatial model mainly extracts the spatial and local temporal features from the frames and temporal model gathers the global temporal semantic features from the output of spatial model.
As temporal model contains less redundancy, \ModelAbbr~is only applied to the spatial model. 

\begin{figure}[t]
    \centering
    \includegraphics[width=\linewidth]{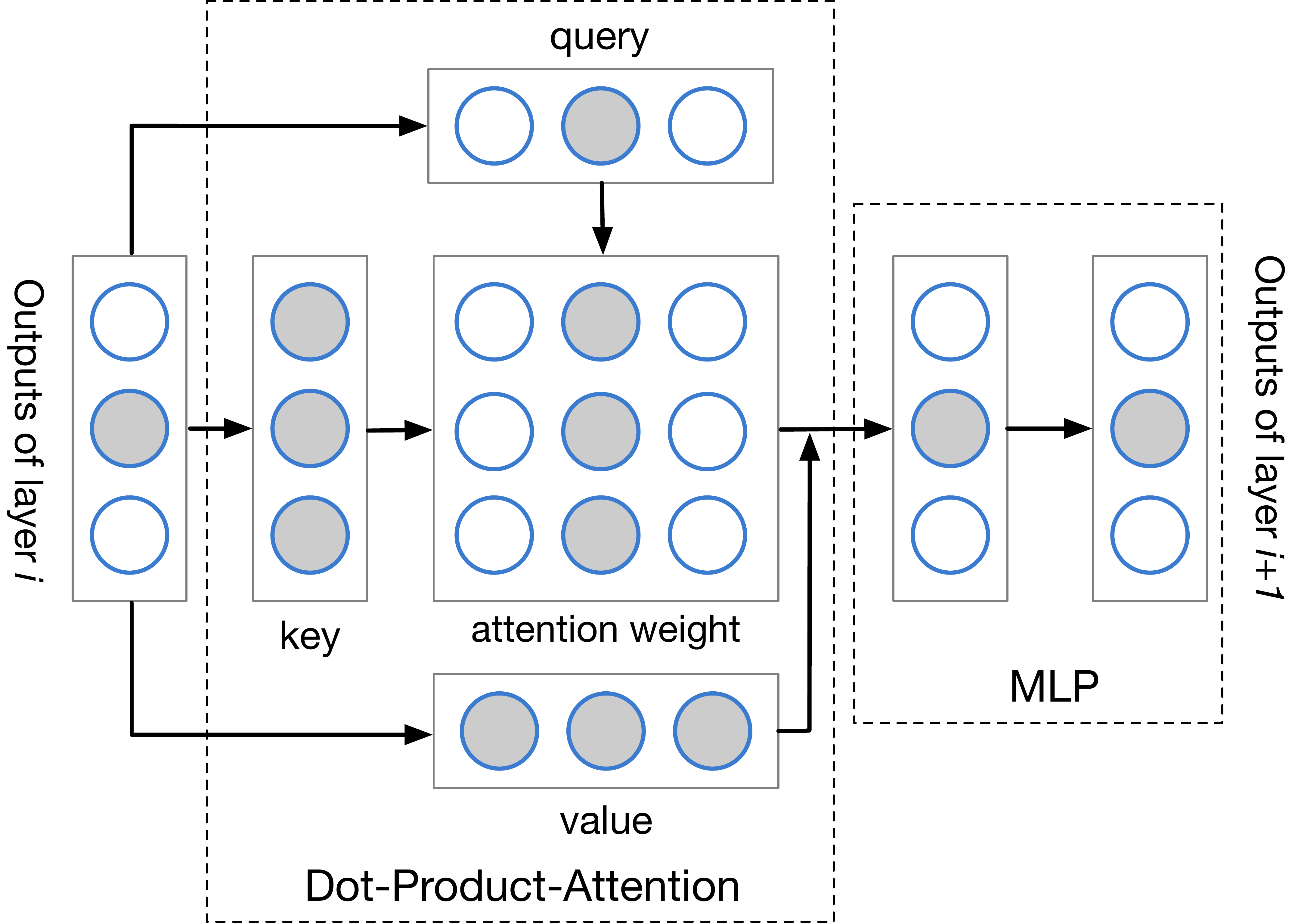}
    \vspace{-3.5mm}
    \caption{\ModelAbbr~on a Dot-Product Attention layer. In this example, the keep-ratio is $1/3$. Only sampled nodes (in color gray) are cached for gradient calculations in model training.}
    \label{fig:sbp_transformer_layer}
    \vspace{-2.5mm}
\end{figure}

\ModelAbbr~can be implemented by dividing the input frames into two sets using a predefined sampling function: Set-1 $\{x_{i_0},x_{i_1},...,x_{i_u}\}$ in which the gradients are kept and Set-2 $\{x_{j_0},x_{j_1},...,x_{j_v}\}$ in which the gradients are dropped. During forward, we track all the activations for Set-1 and only track the extracted features for Set-2. Then we concatenate the extracted features and do fully forward and backward on the temporal model $f_t$.

\vspace{-2.5mm}
\subsubsection{Video Transformers}
\vspace{-1.0mm}

For short-term video tasks, such as action recognition, video Transformers such as TimeSformer~\cite{bertasius2021space} and Video Swin Transformer~\cite{liu2021video} achieve superior accuracy. Here we instantiate \ModelAbbr~on Video Transformers.

In the tree-models, the input node at one temporal location is independent of the nodes at other temporal locations in the spatial model, and thus we can easily apply \ModelAbbr~on it. However, in video Transformers, the nodes at different temporal locations are dependent on each other due to the dot-product attention layer. We address this issue by applying \ModelAbbr~layer-wise instead of model-wise in tree-models.

Fig.~\ref{fig:sbp_transformer_layer} shows how \ModelAbbr~is applied to a Transformer layer. The same set of nodes are sampled (gray nodes in Fig.~\ref{fig:sbp_transformer_layer}) for all the layers. In the Dot-Product-Attention, all the key and value nodes are related to each of the query nodes. Given the gradients of the sampled nodes in the output, in order to calculate the gradients of the corresponding nodes in query accurately, all the key and value nodes are sampled. Sampling all the key and value nodes will only increase the memory consumption slightly as the attention weights comprise most of the GPU memory usage in Dot-Product-Attention.
Suppose the shape of query, key and value is $\mathbb{R}^{ (h \times d) \times n}$ and the shape of attention weights is $\mathbb{R}^{h \times n \times n}$, in which $h$, $d$ and $n$ are the number of heads, the dimension of each head and the number of tokens respectively. 
In video Transformers, the number of tokens $n$ are much larger than the head dimension $d$ (\ie, in video Swin Transformer, $n=392$ and $d=32$) and thus attention weights comprise most of the GPU memory.

In Video Transformers, we consider the several top most layers (3 layers in this paper) served as ``temporal model'' that have fully backward paths and apply \ModelAbbr~only to the bottom layers (spatial model).

\begin{algorithm}[t]
\caption{Pytorch-like pseudocode of \ModelAbbr~for forward and backward on an arbitrary operation $f$.}
\label{alg:sbp_impl}
\definecolor{codeblue}{rgb}{0.25,0.5,0.5}
\lstset{
  backgroundcolor=\color{white},
  basicstyle=\fontsize{8pt}{8pt}\ttfamily\selectfont,
  columns=fullflexible,
  breaklines=true,
  captionpos=b,
  commentstyle=\fontsize{8pt}{8pt}\color{codeblue},
  keywordstyle=\fontsize{8pt}{8pt},
}
\begin{lstlisting}[language=python]
# f: an arbitrary operation
# idx: sampled indices where gradients are kept

# Forward function
def FORWARD(ctx, inputs, idx):
    # forward without gradient calculation
    with torch.no_grad():
        y = f(inputs)
    # cache sampled nodes
    ctx.save_for_backward(inputs[idx])
    # cache sampled index
    ctx.idx = idx
    return y

# Backward function
def BACKWARD(ctx, dy):
    dy = dy[ctx.idx]
    nodes = ctx.saved_tensors 
    with torch.enable_grad():
        # reforward with gradient calculation
        nodes = nodes.detach().requires_grad_(True)
        y = f(nodes)
    d_nodes = torch.autograd.grad(y, nodes, dy)
    # gradients w.r.t inputs
    d_inputs = zeros.fill(ctx.idx, d_nodes)
    return d_inputs
\end{lstlisting}
\end{algorithm}

\subsection{Efficient Implementation}

Alg.~\ref{alg:sbp_impl} shows the implementation for an arbitrary operation $f$, including the spatial model in StT, and the dot-product attention and multilayer perception (MLP) in Video Transformers. During forward, the output is produced without tracking the activations. Only the sampled inputs that need to calculate the gradients are saved for backward. During backward, \ModelAbbr~does the re-forward and backward on the sampled inputs to calculate the gradients. This can take full use of the ``autograd engine'' in the deep learning frameworks and is efficient with little overhead.

\subsection{Space and Time Complexity}

For the StT model, $M_s$ is the memory needed by the spatial model and $M_c$ is the memory needed by the temporal model.
Thus its space ratio is $(rM_s + M_c)/(M_s + M_c)$, which is close to $r$ since $M_s$ is typically 5$\times$ to 10$\times$ larger than $M_c$~\cite{xu2019temporal,xu2021long,xu2020g}.
For Video Transformers, $d$ and $n$ are the head dimensions in multi-head attention, number of tokens, respectively.
The activation maps needed for gradient calculation in a Transformer layer includes 2$\times$ inputs ($2hdn$), QKV vectors (each with $hdn$), the attention weights ($2hnn$ before and after the softmax) and
the MLP\footnote{It consists of norm ($hdn$) $\to$ fc ($4hdn$) $\to$ activation ($4hdn$) $\to$ fc ($hdn$), where the hidden size of fc may vary but is $4hd$ in most vision Transformers~\cite{bertasius2021space,liu2021video}.} ($10hdn$),
thus requires the memory of $15hdn + 2hnn$ in total.
\ModelAbbr~(as shown in Fig.~\ref{fig:sbp_transformer_layer}) reduces the memory of query to $hdnr$, attention weights to $2hnnr$ and MLP to $10hdnr$. So the space ratio is 
\begin{equation}
    \footnotesize
    \frac{4hdn+11hdnr+2hnnr}{15hdn+2hnn} = \frac{r(11\frac{d}{n}+2) +4\frac{d}{n}}{(11\frac{d}{n}+2)+4\frac{d}{n}}
\end{equation}
For both the models, the space-ratio is almost proportional to the sampling rate $r$ and thus the memory reduction of \ModelAbbr~can be significant if $r$ is small (\ie, 0.25). This is because during training, the most memory-consuming part is to cache the activations for gradient calculation, but with \ModelAbbr, we only need to cache the sampled activations.

Regarding the time ratio, suppose the forward and backward time is equivalent, the overall ratio is $\frac{1+2r}{2}$ as we perform $(1+r)\times$ the forward and $r\times$ the backward.

%% file: experiments.tex
\vspace{-1mm}
\section{Experiments on Action Recognition}
\label{sec:exp:actrec}

We evaluate \ModelAbbr~by applying it to state-of-the-art action recognition models, Video Swin~\cite{neimark2021video}, and conduct experiments on Kinetics-400 (K400)~\cite{kay2017kinetics} and EPIC-Kitchens-55 (Epic-55)~\cite{damen2018scaling} datasets.
K400 contains $\sim$240K training and 20K validation videos with 400 actions.
Each video is trimmed to 10 seconds.
Epic-55 includes egocentric videos of unscripted activities (mostly cooking) recorded in kitchen environments. Segments of each video are labeled with one verb and one noun. We follow prior work~\cite{wu2019long} to use their train and validation splits, and report the Top-1 and Top-5 accuracy on verb classification.

\begin{table}[t]
    \centering
    \setlength{\tabcolsep}{4pt}
    \footnotesize
    \begin{tabular}{lcccccc}
        \toprule
        \multirow{2}{*}{Model} & \multirow{2}{*}{Training}  & \multicolumn{2}{c}{K400 (\%)} & \multicolumn{2}{c}{Epic-55 (\%)} & \multirow{2}{*}{\makecell{Mem\\(GB/gpu)}} \\
        \cmidrule(lr){3-4} \cmidrule(lr){5-6}
        & & Top-1 & Top-5 & Top-1& Top-5 & \\
        \midrule
        \multirow{3}{*}{Swin-T} & E2E & 78.8 & 93.6 & 46.0 & 81.5 & 15.2\\ 
        \cmidrule(lr){2-7}
        & FD & 76.3 & 92.4 & 38.9 & 79.9 & \textbf{2.9} \\
        & \ModelAbbr~& \textbf{78.2} & \textbf{93.1} & \textbf{45.2} & \textbf{81.7} & 4.4\\
        \midrule
        \multirow{3}{*}{Swin-B} & E2E & 82.7 & 95.5 & 49.4  & 83.2 & 32.5 \\ 
        \cmidrule(lr){2-7}
        & FD & 80.8 & 94.9 & 44.8 & 82.5 & \textbf{6.3}\\
        & \ModelAbbr~& \textbf{81.9} & \textbf{95.2} & \textbf{47.3} & \textbf{83.6} &  8.6\\
        \bottomrule
    \end{tabular}
    \vspace{-2.5mm}
    \caption{Results of action recognition using Video Swin-T and -B. Both SBP and Frame Dropout (FD) use the keep-ratio at 0.25 here.}
    \vspace{-1.5mm}
    \label{tab:actrec}
\end{table}

\subsection{Implementation Details}
\label{sec:exp:actrec:implementation}

We train Video Swin-T and -B~\cite{neimark2021video} as backbones with batch size 4 on each GPU. 
We use the learning rate of $2.5e$-$4$ and $7.5e$-$5$ for Swin-T and -B respectively and train the models for 30 epochs.
Other hyperparameters are kept the same as the original paper~\cite{neimark2021video}.
During inference, we follow prior work~\cite{wu2019long} by averaging the predictions of $4\times3$ views (\ie, 4 uniform temporal cropping and 3 spatial cropping from top left, center, and bottom right) for K400 and using only central crop for Epic-55.
For \ModelAbbr,~we use the uniform random sampler along the temporal dimension such that the nodes corresponding to one frame are either all sampled or dropped. 
The sampled nodes are kept the same across all layers.
\ModelAbbr~is applied on the first 8 layers for Video Swin-T with layers \{2,2,6,2\}, and is applied on the first 18 layers for Video Swin-B with layers \{2,2,18,2\}.

\begin{table}[t]
    \centering
    \footnotesize
    \setlength{\tabcolsep}{5pt}
    \begin{tabular}{lcccccccc} 
    \toprule
        \multirow{3}{*}{Model} & \multicolumn{4}{c}{Mem (GB/gpu)} & \multicolumn{4}{c}{Speed (s/itr)} \\
        \cmidrule(lr){2-5} \cmidrule(lr){6-9}
        & \multicolumn{2}{c}{without} & \multicolumn{2}{c}{with} & \multicolumn{2}{c}{without} & \multicolumn{2}{c}{with} \\
        \cmidrule(lr){2-3} \cmidrule(lr){4-5} \cmidrule(lr){6-7} \cmidrule(lr){8-9}
        & E2E & \ModelAbbr~& E2E & \ModelAbbr~& E2E & \ModelAbbr~& E2E & \ModelAbbr~\\
        \hline
        Swin-T & 15.2 & 4.4 & 5.8 & \textbf{3.2} & 0.22 & \textbf{0.21} & 0.29 & 0.26 \\
        Swin-B & 32.5 & 8.6 & 8.0 & \textbf{4.6} & 0.43 & \textbf{0.41} & 0.57 & 0.52 \\
    \bottomrule
    \end{tabular}
    \vspace{-2.5mm}
    \caption{Benchmark of the memory usage and training speed of \ModelAbbr~with and without gradient checkpoint. The keep ratio is 0.25 for \ModelAbbr~and batch size is 4 per GPU.
    The memory is measured using \textit{torch.cuda.max\_memory\_allocated()}.
    The speed is measured on a single Nvidia A100 graphics card.}
    \vspace{-1.5mm}
    \label{tab:grad_checkpoint}
\end{table}

\begin{table}[t]
    \centering
    \setlength{\tabcolsep}{5pt}
    \footnotesize
    \begin{tabular}{lccccc}
        \toprule
        \multirow{2}{*}{Training} & \multirow{2}{*}{\makecell{Sampling\\Dimension}} & \multicolumn{2}{c}{K400 (\%)} & \multicolumn{2}{c}{Epic-55 (\%)} \\
        \cmidrule(lr){3-4} \cmidrule(lr){5-6}
        & & Top-1 & Top-5 & Top-1& Top-5 \\
        \midrule
        \multirow{2}{*}{\ModelAbbr} & temporal & 78.2 & \textbf{93.1} & 45.2 & \textbf{81.7}  \\
        &  spatial \& temporal & \textbf{78.3} & 93.0 & \textbf{45.4} & 81.3  \\
        \bottomrule
    \end{tabular}
    \vspace{-2.5mm}
    \caption{Temporal (frame-wise) and spatial-temporal sampling (3D-checkerboard) achieve comparable results.
    The model is Video Swin-T and the keep-ratio of SBP is 0.25.}
    \label{tab:actrec:sampling}
    \vspace{-2.5mm}
\end{table}

\subsection{Evaluation Results}
\label{sec:exp:actrec:results}
\vspace{-0.5mm}

\vspace{-5pt}
\xhdr{Comparison to prior work.}
In Table~\ref{tab:actrec}, we compare the accuracy and memory usage of \ModelAbbr~(w/ keep-ratio as 0.25) to existing training strategies using Video Swin on K400 and Epic-55. Results show that \ModelAbbr~uses only 25\% of memory as end-to-end (E2E) training uses, with only $\sim$0.7\% Top-1 accuracy drop on K400 and $\sim$1.0\% on Epic-55.
Frame Dropout (FD)~\cite{akbari2021vatt} is a simple frame selection method that only processes a subset of uniformly sampled frames.
Different from \ModelAbbr~that only drops the backward paths, FD removes both forward and backward paths for the dropped frames.
Table~\ref{tab:actrec} shows that, although FD costs the least memory, it leads to more significant decrease of accuracy, especially on temporal-heavy dataset, Epic-55.
This demonstrates the importance of keeping all the forward paths during training and inference.

\xhdr{Can \ModelAbbr~work with other efficient training strategies?}
We validate if \ModelAbbr~is complementary with other memory saving methods.
As gradient checkpoint~\cite{chen2016training} is widely used to save memory, we report the memory and speed of \ModelAbbr~when integrated with gradient checkpoint.
As shown in Table~\ref{tab:grad_checkpoint}, the memory consumption of \ModelAbbr~without checkpoint is similar to the standard end-to-end training with checkpoint, but is $\sim$30\% faster. When applying gradient checkpoint to \ModelAbbr,~additional $\sim$40\% the memory can be saved comparing to using \ModelAbbr~alone.
Furthermore, we test to incorporate Mixed Precision (w/ opt\_level as O1) into \ModelAbbr.~The results show that this can lead to $\sim$50\% more savings of GPU memory on Swin-T without accuracy loss.

\xhdr{Can we drop gradients in both spatial and temporal domains?}
\label{sec:exp:actrec:sampling}
Till now, nodes corresponding to one frame are either all kept or dropped.
To verify if \ModelAbbr~can be applied to both spatial and temporal dimensions, we implement the spatial-temporal gradient drop with a 3D-checkerboard pattern. From Table~\ref{tab:actrec:sampling}, temporal sampling (frame-wise) and spatial-temporal (3D-checkerboard) sampling achieve comparable accuracy, which shows \ModelAbbr~also works well along spatial sampling. This result indicates the potential to apply the proposed method to image domains.

\begin{table}[t]
    \centering
    \footnotesize
    \begin{tabular}{lcccc}
        \toprule
        \multirow{2}{*}{Method} & \multirow{2}{*}{Training} & \multirow{2}{*}{mAP (\%)} & \multirow{2}{*}{\makecell{GPU Mem\\(\#GPUs $\times$ GB)}}\\
        & & &\\
        \midrule
        \multirow{5}{*}{TRN~\cite{xu2019temporal}} & Feat. Extract. & 55.3 & $1\times12$ \\
        & E2E & 56.8 & $8\times14.6$ \\
        \cmidrule{2-4}
        & SBP (0.25) & 56.7 & $8\times5.6$ \\
        & SBP (0.125) & \textbf{56.9} & $8\times 4.3$ \\
        \midrule
        \multirow{4}{*}{LSTR~\cite{xu2021long}} & Feat. Extract. & 56.8 & $1\times 5$ \\
        & E2E & 59.2 & $8\times 32$  \\
        \cmidrule{2-4}
        & SBP (0.25) & 59.1  & $8 \times 12.2$  \\
        & SBP (0.125) & \textbf{59.5} & $8\times 7$  \\
        \bottomrule
    \end{tabular}
    \vspace{-2.5mm}
    \caption{Comparison of \ModelAbbr, using a fixed feature extractor (Feat. Extract.), and end-to-end (E2E) training on TRN and LSTR.}
    \label{tab:oad_comp_sota}
    \vspace{-2.5mm}
\end{table}

\vspace{-1.0mm}
\section{Experiments on Temporal Action Detection}
\label{sec:exp:tad}

To validate \ModelAbbr's efficiency for long-term video tasks, we build it upon multiple state-of-the-art methods~\cite{xu2019temporal,xu2021long,xu2020g} for temporal action detection in untrimmed videos.
The experiments are conducted on THUMOS'14~\cite{THUMOS14} dataset, which contains about 20 hours of video with 20 sports actions.
Since its training set contains only trimmed videos that cannot be used to train temporal action
detection models, we follow prior work~\cite{xu2019temporal,xu2021long} and train on the validation set (200 videos) and evaluate on the test set (213 videos). Most experiments are conducted on 8 Tesla 16GB V100, except for the end-to-end training that requires more memory,
the models are trained on 8 Tesla 32GB V100.

\begin{table*}[t]
    \centering
    \footnotesize
    \begin{tabular}{lcccccccccc}
        \toprule
        \multirow{2}{*}{Training} & \multirow{2}{*}{Keep-ratio} & \multicolumn{6}{c}{mAP (\%)} & \multirow{2}{*}{\makecell{GPU Mem\\(\#GPUs $\times$ GB)}}\\
        \cmidrule(lr){3-8}
        & & 0.3 & 0.4 & 0.5 & 0.6 & 0.7 & AVG & \\
        \midrule
        Feat. Extract. & - & 43.15 & 35.84 & 28.36 & 19.89 & 11.96 & 27.84 & \textbf{8$\times$3}\\
        End-to-End & - & 46.14 & \textbf{39.86} & \textbf{32.02} & \textbf{23.29} & \textbf{14.56} & \textbf{31.17} & 8$\times$26 \\
        \midrule
        \multirow{3}{*}{\ModelAbbr}&  0.5 & 45.36 & \textbf{38.72} & \textbf{31.46} & 22.74 & 14.01 & \textbf{30.46} & 8$\times$12 \\
        & 0.25 & 45.12 & 37.80 & 30.44 & \textbf{22.81} & \textbf{14.52} & 30.14 & 8$\times$6 \\
        & 0.125 & \textbf{46.40} & 38.50 & 29.95 & 20.37 & 11.82 & 29.41 & \textbf{8$\times$5}\\
        \bottomrule
    \end{tabular}
    \vspace{-1.5mm}
    \caption{Results of temporal action localization on THUMOS'14. We only use RGB frames as inputs for simplicity. Note that the end-to-
    end training is almost impossible for many research groups but \ModelAbbr~with keep ratio 0.5 can be trained on mainstream GPUs (12 GB).}
    \label{tab:tad_compare}
    \vspace{-0.5mm}
\end{table*}

\vspace{-0.5mm}
\subsection{Online Action Detection}
\label{sec:exp:tad:online}

Given a live video stream, online action detection tries to detect the actions that are occurring in each frame without seeing the future.
Existing methods are formulated in the spatial-then-temporal (StT) paradigm, where the spatial model is pretrained to extract features from the video frames, and the temporal model operates on the feature sequence to capture long-range context.
We adopt two state-of-the-art methods (\ie, TRN~\cite{xu2019temporal} and LSTR~\cite{xu2021long}) and conduct experiments using RGB as inputs\footnote{In the original papers, they ensemble RGB and optical flow as inputs. Without loss of generality, we only use RGB in our experiments.}.
In particular, TRN is built upon recurrent networks for temporal modeling and LSTR uses multiple cascaded Transformers to capture both long- and short-term dependencies.

\vspace{-0.5mm}
\subsubsection{Implementation Details}
\label{sec:exp:tad:online:implementation}
\vspace{-1mm}

We follow the common setting of existing methods~\cite{xu2019temporal,xu2021long}. Specifically, the spatial model is ResNet50 pretrained on K400. The model is trained with batch size 16 and the initial learning rate is $1e$-$6$ and $5e$-$5$ for TRN and LSTR respectively.
Note that both TRN and LSTR freeze their spatial model and only train the temporal model in the original papers.
TRN takes 64 frames as inputs and LSTR takes 160 frames (\ie, long-term 128 and short-term 32) as inputs.
When converting it to end-to-end training using \ModelAbbr,~we set the learning rate of spatial model $\sim$10$\times$ smaller than the initial learning rate (this setting is validated in the later section). As both TRN and LSTR are StT models, we use the implementation as described in Sec.~\ref{sec: sbp_stmodel}.
Per-frame mean average precision (mAP) is adopted to evaluate the performance of online action detection.

\vspace{-3.5mm}
\subsubsection{Evaluation Results}
\vspace{-1mm}

The results are shown in Table~\ref{tab:oad_comp_sota}. First, freezing the backbone leads to 1.5\% and 2.4\% mAP drop for TRN and LSTR respectively. Second, by applying \ModelAbbr~with keep-ratio of 0.125, we can train the model with 22\% of the memory with similar mAPs compared with end-to-end training.
Third, one interesting observation is that \ModelAbbr~with lower keep-ratio $r$=$0.125$ is even slightly higher than the end-to-end training,
which may be the reason that \ModelAbbr~can be served as a regularization to ease overfitting on THUMOS dataset that is relatively small~\cite{tseng2020regularizing}.
Fourth, the improvement on LSTR is larger than that on TRN, which indicates a stronger temporal model can benefit more from the end-to-end training.

\vspace{-2.5mm}
\subsubsection{Ablation Studies}
\vspace{-1mm}

\begin{figure}
    \centering
    \includegraphics[width=\linewidth]{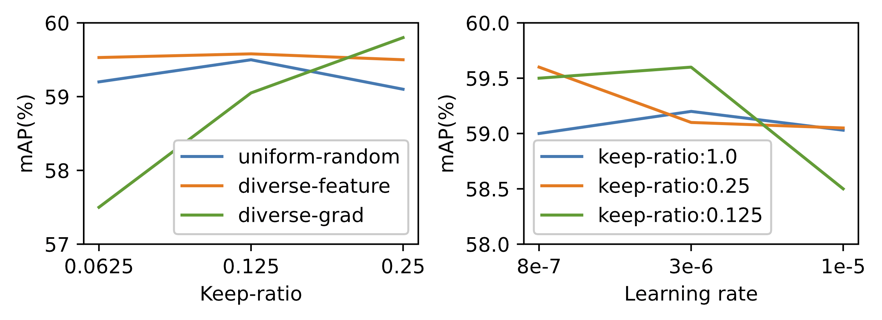}
    \vspace{-7.0mm}
    \caption{The influence to the performance in mAP of using different sampling methods and initial learning rates for spatial model. The initial learning rate for temporal model is 5e-5.}
    \label{fig:ablation_online}
    \vspace{-1.5mm}
\end{figure}

\vspace{-5pt}
\xhdr{How to sample nodes?}
One key aspect of \ModelAbbr~is how to sample the output nodes of the top most layer that \ModelAbbr~is applied to (named as \textit{candidate nodes} for abbreviation).
We examine the three sampling methods as below.
(1) \textit{Uniform Random Sampler} evenly divides the candidates nodes into chunks along temporal dimension. The size of each chunk is $1/r$. We randomly sample 1 node inside each chunk.
(2) \textit{Diverse-Feature Sampler} first sorts the candidate nodes according to the magnitude and then performs uniform random sampling on the sorted node list.
(3) \textit{Diverse-Grad Sampler} sorts the gradients of the candidate nodes according to the magnitude and then performs uniform random sampling on the sorted node list.
From the results in Fig.~\ref{fig:ablation_online}, we can observe that: 1) the diverse-feature sampler achieves overall the best mAP; 2) the mAP of these three samplers are quite similar, which indicates our method is robust to the samplers. As the uniform random sampler achieves comparable mAP with others but is much simpler to implement (preferred from the perspective of Occam’s razor), we adopt it as our default sampler for the following experiments.

\xhdr{How to set learning rate for the spatial model?}
We do some empirical studies to determine the best learning rate for training StT models with \ModelAbbr.~The results are shown in Fig.~\ref{fig:ablation_online}. We observe that: 1) the best learning rate for spatial model is about $10\times$ smaller than the temporal model for all the keep-ratios; 2) training with \ModelAbbr,~the spatial model is quite robust to the learning rate as long as they are relative small. Therefore, unless noted otherwise, we set the learning rate of spatial model as $0.1\times$ of the temporal model.

\subsection{Temporal Action Localization}
\label{sec:exp:tad:offline}

Temporal Action localization aims to detect the temporal boundaries for every action instance in untrimmed videos. Most existing methods~\cite{gao2017turn,lin2019bmn,xu2020g} follow the StT paradigm. We take the state-of-the-art method, G-TAD~\cite{xu2020g} as baseline and conduct experiments on THUMOS'14~\cite{idrees2017thumos}.

\vspace{-3.5mm}
\subsubsection{Implementation Details}
\label{sec:exp:tad:offline:implementation}
\vspace{-1.5mm}

We use the same settings as in the original paper~\cite{xu2020g}: the spatial model is ResNet50 pretrained on K400 and the learning rate for temporal model is $4e$-$5$. When integrating it with \ModelAbbr,~we unfreeze the spatial model and set the learning rate for spatial model to $5e$-$5$.
We take mean Average Precision (mAP) at certain IoUs as the evaluation metric. 

\vspace{-3.5mm}
\subsubsection{Evaluation Results}
\label{sec:exp:tad:offline:results}
\vspace{-1.5mm}

The results are as shown in Tab~\ref{tab:tad_compare}. Freezing the backbone leads to 3.3\% drop in average mAP. \ModelAbbr~is $\sim$1\% lower than end-to-end training but can still lead to $\sim$2\% improvement comparing to using frozen feature extractor. Training with \ModelAbbr~uses only 12 GB, 6 GB and 5 GB on each GPU compared with 26GB using end-to-end training for keep-ratio of 0.5, 0.25 and 0.125 respectively. Note the amount of memory (8 GPUs, each with 26GB) for end-to-end training may not be available for many research groups. But applying \ModelAbbr~with keep-ratio $0.25$ only gets $1\%$ loss of mAP, the model can be trained on 4 GPUs with 12 GB memory (\ie, $4\times$ GTX 1080Ti), which is much more feasible.

%% file: conclusion.tex
\section{Discussion}

We presented a simple yet effective technique \ModelName~(\ModelAbbr)~that can reduce a large portion of GPU memory usage for training video models.
However, there are still several limitations of our current work.
First, \ModelAbbr~is a new prototype to show that memory saving can be achieved by dropping the backward paths, but as byproduct, it still causes slightly loss of accuracy. Future explorations on strategies of using more adaptive layer-wise keep-ratios and sampling methods can be studied to reduce the loss of accuracy or even improve the accuracy due to the regularization effect of gradient dropout~\cite{tseng2020regularizing}.
Second, \ModelAbbr~brings only a small amount of training speedup ($\sim$1.1$\times$). However, how to integrate it with other efficient model training strategies, such as Multigrid~\cite{wu2020multigrid}, to speed up the training is still not well explored.

\xhdr{Potential Negative Impact}.
We study the research problem of efficient training for video models, which can be used in real-world video understanding applications, such as patient or elderly health monitoring, augmented and virtual reality, and collaborative robots.
But there still could be unintended uses, and we advocate responsible usage complying applicable laws and regulations.